%% file: main.tex
\documentclass[10pt,twocolumn,letterpaper]{article}

\usepackage{cvpr}              

\makeatletter
\newcommand{\paragrapht}{\@startsection{paragraph}{4}{\z@}%
  {0pt}
  {-1em}
  {\normalfont\normalsize\bfseries}}
\makeatother

\input{preamble}

\definecolor{cvprblue}{rgb}{0.21,0.49,0.74}
\usepackage[pagebackref,breaklinks,colorlinks,allcolors=cvprblue]{hyperref}

\usepackage{multirow}
\usepackage{gensymb}
\usepackage{comment}
\usepackage{graphicx}
\usepackage{amsmath, amssymb}
\usepackage{bbm}

\usepackage{graphicx}  

\usepackage{makecell}

\usepackage{caption}
\usepackage{booktabs}           
\usepackage{float}
\usepackage{amsfonts}
\usepackage{xcolor}
\usepackage{mathtools}
\usepackage{algorithm}
\usepackage[noend]{algpseudocode}

\newcommand{\ours}{Refin3R}

\newcommand{\res}[2]{#1/#2}

\def\confName{CVPR}
\def\confYear{2027}

\title{Refin3R: Geometry-Guided Refinement for Feed-Forward 3D Reconstruction}

\author{
Jongmin Lee \quad
Jaeyeong Kim \quad
Seungryong Kim\textsuperscript{\(\dagger\)}\\[2mm]
KAIST AI\\[1mm]
{\tt \url{https://cvlab-kaist.github.io/Refin3R}}
}

\begin{document}

\input{figures/teaser}

\renewcommand{\thefootnote}{\fnsymbol{footnote}}
\footnotetext[2]{Corresponding author.}
\renewcommand{\thefootnote}{\arabic{footnote}}

\input{sec/0_abstract}    
\vspace{-5mm}
\input{sec/1_intro}

\input{sec/2_related_works}

\input{sec/3_method}

\input{sec/4_experiment}

\input{sec/5_conclusion}
{
    \small
    \bibliographystyle{ieeenat_fullname}
    \bibliography{main}
}


\end{document}

%% file: preamble.tex
\usepackage[table]{xcolor}   

\definecolor{idcolor}{HTML}{DCEBFA}   
\definecolor{cdcolor}{HTML}{E3F4E1}   
\definecolor{odcolor}{HTML}{FBE5D6}   

\newcommand{\dshl}[2]{{\setlength{\fboxsep}{1pt}\colorbox{#1}{\strut #2}}}
\newcommand{\dsid}[1]{\dshl{idcolor}{#1}}
\newcommand{\dscd}[1]{\dshl{cdcolor}{#1}}
\newcommand{\dsod}[1]{\dshl{odcolor}{#1}}



%% file: figures/teaser.tex

\twocolumn[{%
  \maketitle
  \begin{center}
    \includegraphics[width=\textwidth]{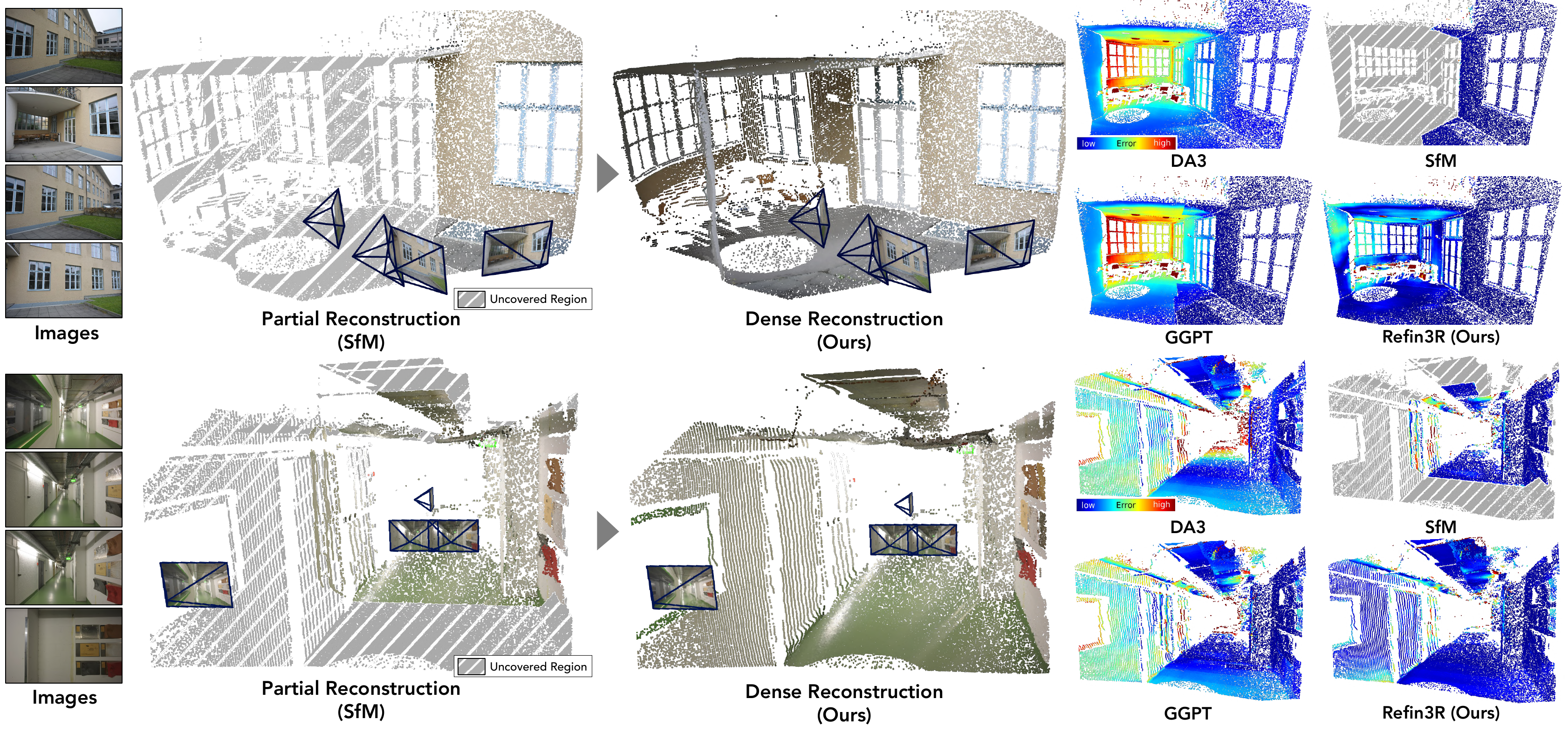}
    \vspace{-2em}
    \captionof{figure}{\textbf{\ours{} refines feed-forward 3D reconstruction with SfM points.}
    Given multi-view images, feed-forward models such as Depth Anything 3 (DA3)~\cite{lin2025depth} predict dense but less accurate point maps, while SfM points are accurate but incomplete.
    Leveraging both, \textbf{\ours{}} produces dense and accurate point maps.
    \textbf{Right}: error maps of DA3, SfM (GGPT pipeline~\cite{chen2026ggpt}), GGPT, and \ours{}.    
    } \label{fig:teaser}
  \end{center}
}]

%% file: sec/0_abstract.tex
\begin{abstract}

Feed-forward 3D reconstruction models predict dense point maps from multiple images in a single pass. Without explicit geometric constraints, however, their predictions are often inconsistent across views and lack fine-grained precision. 
Recent methods incorporate geometric priors such as camera pose or depth, but use them merely as auxiliary inputs, process each view independently, or refine the geometry only where the priors are available.
We present \ours{}, which refines feed-forward point maps with Structure-from-Motion (SfM) points, propagating SfM geometry over the image grid.
In its dual-branch architecture, a 2D branch extends SfM geometry beyond its coverage using geometry-aware backbone features, while a 3D branch reasons over the SfM points of all views, capturing multi-view structure that the image grid alone cannot represent. The two branches exchange information bidirectionally to jointly refine each view, yielding dense and accurate point maps.
Across ScanNet++, ETH3D, Tanks\&Temples, and out-of-domain human and surgical scenes, \ours{} consistently outperforms feed-forward and geometry-guided methods in point map accuracy while running faster than the strongest geometry-guided baseline.

\end{abstract}

%% file: sec/1_intro.tex
\section{Introduction}
\label{sec:intro}

Multi-view 3D reconstruction from RGB images is a long-standing problem in computer vision.
Recent feed-forward methods~\cite{wang2024dust3r,wang2025vggt,lin2025depth}, trained on large-scale datasets, predict dense point maps for all views in a single forward pass. Their fast and scalable inference has established this paradigm as a promising direction, and follow-up works are rapidly emerging~\cite{wang2026pi, keetha2026mapanything, wang2026vggt}. However, these models regress point maps purely from learned priors, without enforcing explicit geometric constraints such as triangulation~\cite{hartley1997triangulation}. As a result, their predictions are often inconsistent across views and lack fine-grained precision.

Structure-from-Motion (SfM)~\cite{colmap, sfm1, incrementalsfm, build_rome_in_a_day}, in contrast, establishes correspondences across images and triangulates them under explicit multi-view constraints, yielding consistent and accurate points. Yet it often recovers only sparse geometry and becomes fragile under wide baselines or low overlap~\cite{schonberger2016pixelwise, pataki2025mp}. Moreover, since it relies solely on correspondences without semantic understanding, mismatches can introduce outliers~\cite{cai2023doppelgangers}. The two paradigms are thus complementary: feed-forward models are dense but imprecise, whereas SfM is accurate but incomplete.

Recent works incorporate explicit geometric cues, such as correspondences, camera poses, or depth, into feed-forward reconstruction. 
Optimization-based methods~\cite{duisterhof2025mast3r,blast3r} match pixels across views with learned matchers and jointly refine cameras and dense depth, initialized from feed-forward predictions. However, they adjust depth only via a coarse grid of anchor depths or a few per-image coefficients, limiting fine-grained refinement.
Learning-based methods instead feed geometric priors into networks: geometry-conditioned feed-forward models~\cite{jang2025pow3r,keetha2026mapanything} take cameras or depth as extra inputs, and depth completion methods~\cite{murre,omnidc,any2full} densify sparse depth under image guidance. However, the former use the priors merely as auxiliary inputs, while the latter process each view independently, ignoring the 3D structure shared across views.

Closer to our work, GGPT~\cite{chen2026ggpt} refines the feed-forward point maps of all views jointly in 3D space with accurate points from its SfM pipeline, enforcing multi-view consistency around these points.
However, it refines local cubic patches of points from geometric inputs alone, so its refinement remains largely confined to the regions covered by SfM.
Areas without such coverage, typically textureless surfaces and regions observed by a single view, thus benefit little from the SfM guidance (Fig.~\ref{fig:teaser}).

Motivated by these limitations, we propose \textbf{\ours{}}, which refines the feed-forward point maps as per-view depth on the image grid rather than as points in 3D space.
Operating on the image grid allows \ours{} to leverage the dense image features of the pretrained feed-forward backbone~\cite{lin2025depth,wang2026pi}, which guide the refinement across the whole image.
In addition, \ours{} exploits the SfM points of all views in 3D, so that each view is refined with the multi-view structure of the scene rather than only the points it observes.
By combining both, \ours{} recovers accurate point maps even in regions without SfM coverage (Fig.~\ref{fig:teaser}).

We realize this with a dual-branch architecture: a 2D branch on the image grid and a 3D branch on sparse voxels.
The 2D branch refines the feed-forward depth of each view with its own SfM depth, guided by the dense image features from the feed-forward backbone, whose semantic and geometric cues let the SfM depth propagate beyond its coverage. This per-view input, however, holds at most one depth per pixel, so SfM points occluded in the current view or triangulated only from other views are left out.
The 3D branch complements it by gathering the SfM points of all views into sparse voxels within the space seen by each view, providing a richer 3D structure than the view's own SfM depth.
As each voxel is indexed by a pixel and a depth bin, the two branches share the same pixel coordinates and exchange information: the 2D branch receives the multi-view evidence that the image grid alone cannot represent, and the 3D branch receives the image context around each point. 
Spatial propagation then fuses their predictions into the refined depth, which is back-projected to obtain the refined point maps.

Extensive experiments show that \ours{} achieves state-of-the-art point map accuracy on both standard and out-of-domain benchmarks, consistently outperforming feed-forward and geometry-guided methods by a large margin. Moreover, while sharing its SfM front-end with the strongest geometry-guided baseline, \ours{} runs substantially faster, making it an efficient refinement module for feed-forward 3D reconstruction.

%% file: sec/2_related_works.tex
\section{Related works}

\subsection{3D reconstruction}

Multi-view 3D reconstruction has long been studied in computer vision, with its foundations in classical multi-view geometry~\cite{sfm1,sfm2,sfm3,multi_view_geometry}, and has matured into practical SfM systems~\cite{build_rome_in_a_day,incrementalsfm,reconstruct_world,pan2024global} such as COLMAP~\cite{colmap}. 
Learning-based components have since been integrated throughout the SfM pipeline, from feature detection and matching~\cite{detone2018superpoint,superglue,lightglue,sun2021loftr,edstedt2023dkm,edstedt2024roma,edstedt2026roma,ufm,lee2026mv} to track refinement~\cite{pixsfm,dfsfm,lee2025dense}. Beyond individual components, \mbox{VGGSfM}~\cite{wang2024vggsfm} makes the entire pipeline end-to-end differentiable, while MP-SfM~\cite{pataki2025mp} leverages monocular priors to improve robustness under low overlap.
Recently, feed-forward reconstruction models infer scene geometry and camera parameters directly from images. Pairwise models~\cite{wang2024dust3r,mast3r} regress point maps from image pairs but require a global alignment for more views, whereas subsequent multi-view models~\cite{wang2025vggt,wang2026vggt,wang2026pi,lin2025depth,keetha2026mapanything} process all views jointly in a single forward pass. Trained on large-scale 3D data, these models are fast and robust in sparse-view settings, yet without explicit geometric constraints their predictions often lack multi-view consistency and local precision, especially on out-of-domain scenes.

\input{figures/overall_architecture}

\subsection{Geometry-guided reconstruction} 
Recent works leverage both feed-forward models and multi-view geometry, following two directions. One direction refines feed-forward predictions by correspondence-based optimization. VGGT+BA~\cite{wang2025vggt} refines the predicted cameras through bundle adjustment~\cite{bundle_adjustment} to improve their accuracy, and MASt3R-SfM~\cite{duisterhof2025mast3r} and BLASt3R~\cite{blast3r} further optimize dense depth jointly with the cameras.
The other direction feeds geometric priors into feed-forward models, \eg, multi-view transformers that take camera parameters or depth as optional inputs~\cite{jang2025pow3r,keetha2026mapanything, wang2026pi}.
Most closely related, GGPT~\cite{chen2026ggpt} combines both directions: it obtains accurate SfM points with dense matching, sparse bundle adjustment initialized by feed-forward predictions, and linear triangulation~\cite{multi_view_geometry}, and refines the feed-forward point maps with them using a 3D point transformer~\cite{wu2024point}. 
We adopt its SfM pipeline but refine depth in the image domain with geometry-aware image features, while aggregating the SfM points of all views in 3D.

\subsection{Depth completion}

Geometry-guided reconstruction is closely related to depth completion, which densifies sparse depth, typically from LiDAR~\cite{zhang2023completionformer,spnet,wang2025pacgdc,omnidc}, under image guidance, with recent methods leveraging monocular depth foundation models~\cite{any2full,priorda}.
Following this formulation, Murre~\cite{murre} performs multi-view reconstruction by guiding a monocular diffusion model with the SfM depth of each view. 
However, these methods process each view independently and only use the measurements observed in that view. Our image-domain refinement shares this spirit, but both the image features and the SfM evidence in our method are multi-view: the features come from a multi-view feed-forward model, and the 3D branch aggregates the SfM points of all views.

%% file: figures/overall_architecture.tex
\begin{figure*}[t]
    \centering
    \includegraphics[width=\linewidth]{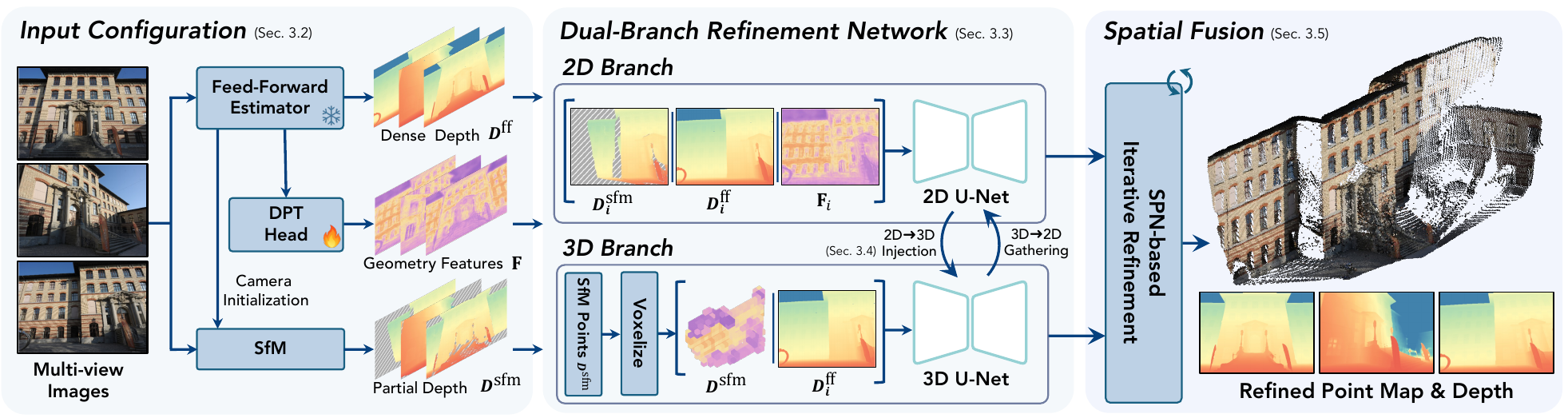}
    \vspace{-4mm}
    \caption{\textbf{Overview of \ours{}.}
    A feed-forward model provides dense depth $D^{\mathrm{ff}}$ and geometry-aware features $\mathbf{F}$, and SfM provides partial but accurate depth $D^{\mathrm{sfm}}$ (Sec.~\ref{sec:method-formulation}).
    For each view, a 2D U-Net refines the dense depth on the image grid, while a sparse 3D U-Net refines the SfM points of all views that project into its image, with both branches exchanging information (Secs.~\ref{sec:method-arch} and~\ref{sec:method-interact}). A spatial propagation network (SPN) fuses their estimates into the refined depth (Sec.~\ref{sec:method-fusion}), which is back-projected into the final point map.}
    \label{fig:overview_arch}
    \vspace{-3mm}
\end{figure*}

%% file: sec/3_method.tex
\section{Method}
\label{sec:method}

\subsection{Overview}

Given $N$ RGB images $\{I_i\}_{i=1}^{N}$, our goal is to reconstruct a dense point map $P_i \in \mathbb{R}^{H\times W\times 3}$ for each view in a shared coordinate system. As illustrated in Fig.~\ref{fig:overview_arch}, \ours{} first obtains two complementary reconstructions: a dense but less accurate one from a feed-forward model and a partial but accurate one from SfM.
Representing both as pixel-aligned depth maps of each view (Sec.~\ref{sec:method-formulation}), it then refines the feed-forward depth on the image grid with per-pixel residuals guided by the SfM points.

The refinement is performed by a dual-branch network that takes both reconstructions along with geometry-aware features from the feed-forward backbone (Sec.~\ref{sec:method-arch}).
Its 2D branch combines the feed-forward and SfM depth of each view with these features and refines the dense depth on the image grid.
In parallel, its 3D branch refines the depth at the SfM-observed pixels of the current view, leveraging the SfM points aggregated from all views in sparse voxels.
The two branches exchange information (Sec.~\ref{sec:method-interact}), and a spatial fusion stage merges their estimates across the image through iterative spatial propagation (Sec.~\ref{sec:method-fusion}). Finally, the refined depth $D^{pred}_i$ is back-projected along the pixel rays of the SfM cameras to obtain the point map $P_i$ of each view.

\subsection{Input representation}
\label{sec:method-formulation}

\paragrapht{Initial reconstructions.}
Following GGPT~\cite{chen2026ggpt}, we obtain two initial reconstructions. A pretrained multi-view feed-forward model, such as Depth Anything 3 (DA3)~\cite{lin2025depth} or Pi3X~\cite{wang2026pi}, first predicts initial camera parameters and a dense depth map $D^{\mathrm{ff}}_i\in\mathbb{R}^{H\times W}$ for every view. Building on this prediction, an SfM pipeline establishes correspondences with RoMa~v2~\cite{edstedt2026roma}, refines the cameras $(\mathbf{K}_i,\mathbf{R}_i,\mathbf{t}_i)$ by sparse bundle adjustment, and triangulates valid tracks by multi-view DLT~\cite{hartley1997triangulation}; see GGPT~\cite{chen2026ggpt} for details. This yields a partial SfM depth map $D^{\mathrm{sfm}}_i$ with a validity mask $M_i\in\{0,1\}^{H\times W}$ marking triangulated pixels.

\paragrapht{Geometric inputs.}
We express both reconstructions as depth along the pixel rays, so that they are aligned on the image grid of each view, and parameterize depth in log space, $\zeta=\log D$, which compresses its wide dynamic range~\cite{moge3}.
Each pixel $p=(u,v)$ of view $i$ is then described by its ray $\mathbf{r}_i(p)=\mathbf{K}_i^{-1}[u,v,1]^\top$, the feed-forward and SfM log-depths $\zeta^{\mathrm{ff}}_i(p)$ and $\zeta^{\mathrm{sfm}}_i(p)$, and the mask $M_i(p)$, where we set $\zeta^{\mathrm{sfm}}_i(p)=0$ if $M_i(p)=0$. 

\paragrapht{Geometry-aware image features.}

As in depth completion, propagating sparse geometry across an image requires dense geometric context, indicating how depth relates between neighboring pixels.
We therefore reuse intermediate features of the frozen feed-forward backbone, whose multi-view transformer is trained to predict geometry and thus carries both semantic and geometric cues.
We take the image tokens $\mathbf{T}\in\mathbb{R}^{4\times(H_p W_p)\times C_t}$ from four intermediate layers of the backbone, where $H_p W_p$ is the number of patches and $C_t$ the token dimension. A DPT head~\cite{ranftl2021vision}, trained end-to-end as part of the full model, fuses and upsamples them into an image-resolution feature map $\mathbf{F}$:

\begin{equation}
  \mathbf{F}=\mathrm{DPT}_{\mathrm{feat}}(\mathbf{T})\in\mathbb{R}^{H\times W\times C_F}.
  \label{eq:feat}
\end{equation}

\subsection{Dual-branch refinement network}
\label{sec:method-arch}

Our setting resembles depth completion, which densifies sparse depth measurements under RGB guidance~\cite{any2full,omnidc,murre}, but differs in two aspects: the feed-forward depth already provides a dense initial estimate, and the partial measurements come from all views rather than the current one alone. We therefore design one branch for each: a dense 2D branch that refines the feed-forward depth on the image grid, and a sparse 3D branch that refines the SfM depth using the points of all views. 
Both predict residuals in log-depth space, the 2D branch on $\zeta^{\mathrm{ff}}_i$  and the 3D branch on $\zeta^{\mathrm{sfm}}_i$ at the SfM-observed pixels.

\paragrapht{2D branch.}
Guided by the image features $\mathbf{F}$, the 2D branch $U_{\mathrm{2D}}$ propagates the partial SfM depth to pixels without any triangulated point, \eg, textureless regions or regions observed by a single view.
It is a U-Net~\cite{unet} built from ConvNeXt blocks~\cite{woo2023convnext} with six scales (stride $1$ to $32$).
Per pixel, it takes the geometric channels:
\begin{equation}
  \mathbf{g}_i(p)=\big[\,\mathbf{r}_i \,\|\, \zeta^{\mathrm{ff}}_i \,\|\, \zeta^{\mathrm{sfm}}_i \,\|\, (\zeta^{\mathrm{sfm}}_i-\zeta^{\mathrm{ff}}_i) \,\|\, M_i\,\big](p),
  \label{eq:geo}
\end{equation}
where $\|$ denotes channel-wise concatenation. 
The difference $\zeta^{\text{sfm}}_i - \zeta^{\text{ff}}_i$ (set to zero where $M_i=0$) makes the discrepancy between the two reconstructions explicit, indicating how much the feed-forward depth deviates from the SfM depth.

Given $\mathbf{g}_i$ and the image features $\mathbf{F}_i$, $U_{\mathrm{2D}}$ predicts a log-depth residual $\Delta\zeta^{\mathrm{2D}}_i$ and a confidence logit $\hat{c}_i$:
\begin{equation}
  (\Delta\zeta^{\mathrm{2D}}_i,\hat{c}_i)=U_{\mathrm{2D}}\big([\,\mathbf{g}_i \,\|\, \mathbf{F}_i\,]\big),\quad
  \zeta^{\mathrm{2D}}_i=\zeta^{\mathrm{ff}}_i+\Delta\zeta^{\mathrm{2D}}_i,
  \label{eq:2d}
\end{equation}
where $\hat{c}_i$ is used for training supervision (Sec.~\ref{sec:method-training}).

\paragrapht{3D branch.}
The 2D branch takes only the SfM depth of the current view, which holds at most one value per pixel; points occluded in this view or triangulated only from other views are thus left out, leaving the full scene geometry unexploited. 
Inspired by multi-view reconstruction methods that aggregate evidence in 3D space~\cite{murez2020atlas,rich20213dvnet}, we introduce a 3D branch that reasons over the SfM points of all views with sparse 3D convolution to refine each view's depth.

Following MoGe-3~\cite{moge3}, we construct for each view $i$ a voxel grid in its image coordinates and log depth at the original image resolution, so that a voxel $(u,v,z)$ corresponds to pixel $(u,v)$ and $z$ indexes its discretized log depth.
We then voxelize the SfM points of \emph{all} views into this grid, keeping only those that lie in front of camera~$i$ and project inside its image.
Further details, including how points are assigned to voxels, are in the supplementary material.

The 3D branch $U_{\mathrm{3D}}$ is a sparse U-Net with five levels of residual sparse 3D convolution blocks on this grid, through which each voxel aggregates features from points close to it in 3D space.
For each occupied voxel $m$, let $p_m$ denote its pixel and $\zeta_m$ the log depth of its SfM point in view $i$.
As in Eq.~\eqref{eq:geo}, $U_{\mathrm{3D}}$ takes per voxel an input feature that combines $\zeta_m$ with the offsets of $\zeta_m$ from the feed-forward and SfM depths at $p_m$:
\begin{equation}
\begin{split}
  \mathbf{v}^{(0)}_m = \big[\,
    &\mathbf{r}_i(p_m)\;\|\;\zeta_m\;\|\;\zeta_m-\zeta^{\mathrm{ff}}_i(p_m)\;\|\; \\
    &\zeta_m-\zeta^{\mathrm{sfm}}_i(p_m)\;\|\;M_i(p_m)\;\|\;\mathbbm{1}^{\mathrm{own}}_m\,\big],
\end{split}
\label{eq:voxinit}
\end{equation}
where $\mathbbm{1}^{\mathrm{own}}_m$ indicates whether $m$ holds a point triangulated by view $i$.
The two differences encode the depth of each SfM point relative to the feed-forward depth and to the pixel's own SfM depth, respectively (the latter difference is set to zero where $M_i(p_m)=0$), so that the scene geometry aggregated from all views is represented in terms of the current view.

From these features, $U_{\mathrm{3D}}$ predicts a log-depth residual and a confidence logit for every occupied voxel at the input resolution, \ie, that of $\mathbf{v}^{(0)}_m$.
Reading them out at the voxels with $\mathbbm{1}^{\mathrm{own}}_m=1$ yields per-pixel maps on $M_i=1$:
\begin{equation}
  (\Delta\zeta^{\mathrm{3D}}_i,\hat{\beta}_i)=U_{\mathrm{3D}}\big(\{\mathbf{v}^{(0)}_m\}_m\big),\quad
  \zeta^{\mathrm{3D}}_i=\zeta^{\mathrm{sfm}}_i+\Delta\zeta^{\mathrm{3D}}_i.
  \label{eq:3d}
\end{equation}
The confidence $\beta_i=\sigma(\hat{\beta}_i)$, where $\sigma$ is the sigmoid function, is used in Sec.~\ref{sec:method-fusion}.

\subsection{Bidirectional 2D--3D interaction}
\label{sec:method-interact}

The two branches are complementary: the 2D branch uses image features to propagate depth to pixels without SfM points, whereas the 3D branch holds multi-view evidence that the image grid alone cannot represent, such as SfM points triangulated without correspondences in view~$i$.
We let them exchange information in their encoders at two coarse scales, $l\in\{3,4\}$ (stride $8$ and $16$), where both branches have the same feature width.
Denoting by $\mathbf{X}^{(l)}_i$ the 2D feature map of view $i$ and by $\mathbf{V}^{(l)}_i$ the voxel features of its grid at scale $l$, each exchange proceeds as:

\begin{equation}
  \begin{aligned}
    \mathbf{X}^{(l)} &\leftarrow \mathrm{Enc}^{(l)}_{\mathrm{2D}}\big(\mathbf{X}^{(l-1)}\big), \\
    \mathbf{V}^{(l)} &\leftarrow \mathrm{Blocks}^{(l)}_{\mathrm{3D}}\big(\mathrm{Inject}_l(\mathbf{V}^{(l)},\mathbf{X}^{(l)})\big), \\
    \mathbf{X}^{(l)} &\leftarrow \mathbf{X}^{(l)} + \mathrm{Gather}_l(\mathbf{X}^{(l)},\mathbf{V}^{(l)}),
  \end{aligned}
  \label{eq:lockstep}
\end{equation}

where $\mathrm{Enc}^{(l)}_{\mathrm{2D}}$ is the scale-$l$ encoder stage of $U_{\mathrm{2D}}$, $\mathrm{Blocks}^{(l)}_{\mathrm{3D}}$ are the sparse convolution blocks of $U_{\mathrm{3D}}$ at the same scale, and $\mathrm{Inject}_l$ and $\mathrm{Gather}_l$ are the two exchange operations detailed below.

Since both branches halve the image axes at each scale, the 2D feature map and the voxel grid at scale~$l$ share the same image grid, downsampled by $2^l$.
Every voxel $m$ thus lies at a pixel $p_m$ of this grid, and we call the set of voxels corresponding to a pixel $p_m$ its \emph{column}.
In the 2D$\rightarrow$3D direction ($\mathrm{Inject}$), voxels in a column receive the image feature at its pixel, $\mathbf{X}^{(l)}_i(p_m)$, by a direct lookup.
In the 3D$\rightarrow$2D direction ($\mathrm{Gather}$), $\mathbf{X}^{(l)}_i(p)$ receives a single feature reduced from the voxel features in its column.

\input{figures/subfigure}

\paragrapht{2D$\rightarrow$3D injection.}
Each voxel feature $\mathbf{v}^{(l)}_m$ is concatenated with the image feature at the corresponding pixel, $\mathbf{X}^{(l)}_i(p_m)$, providing image context absent from the 3D branch's input:
\begin{equation}
  \mathbf{v}^{(l)}_m \leftarrow  \operatorname{MLP} \big([\,\mathbf{v}^{(l)}_m \,\|\, \,\mathbf{X}^{(l)}_i(p_m)\,]\big).
  \label{eq:inject}
\end{equation}
All voxel features in a pixel column receive the same image feature (Fig.~\ref{fig:2d3dinteraction} top).

\paragrapht{3D$\rightarrow$2D gathering.}
The voxel features contain multi-view 3D information, aggregated from the SfM points of all views.
We provide this information to the 2D branch by reducing each column to a single feature at a pixel (Fig.~\ref{fig:2d3dinteraction} bottom).
Since the 2D branch operates on the image of view~$i$, we use only the voxels relevant to this view.
Specifically, we extend $\mathbbm{1}^{\mathrm{own}}$ to scale~$l$ by marking a voxel as own if it contains at least one own voxel at the input resolution (blue in Fig.~\ref{fig:2d3dinteraction} bottom), and exclude voxels that contain only other views' voxels (red in Fig.~\ref{fig:2d3dinteraction} bottom).

The own voxels in each column are then average-pooled along the depth axis:
\begin{equation}
  \mathbf{q}_{p}=\tfrac{1}{|\mathcal{Z}_{p}|}\sum_{m\in\mathcal{Z}_{p}}\bar{\mathbf{v}}^{(l)}_{m},
  \quad
  \mathcal{Z}_{p}=\{m: p_m=p,\ \mathbbm{1}^{\mathrm{own}}_m=1\},
  \label{eq:cellz}
\end{equation}
where $\bar{\mathbf{v}}$ denotes layer-normalized voxel features, and $\mathbf{q}_{p}=\mathbf{0}$ if $\mathcal{Z}_{p}$ is empty.
The pooled feature $\mathbf{q}_{p}$ is then concatenated with the image feature $\mathbf{X}^{(l)}_i(p)$, fed into an MLP, and added to it as a residual:
\begin{equation}
   \mathbf{X}^{(l)}_i(p) \leftarrow \mathbf{X}^{(l)}_i(p) + \operatorname{MLP}\big([\,\mathbf{X}^{(l)}_i(p) \,\|\, \mathbf{q}_{p}\,]\big).
  \label{eq:gather}
\end{equation}

\subsection{Spatial fusion}
\label{sec:method-fusion}

The two branches provide two refined depth estimates: a dense one from the 2D branch and, on SfM-observed pixels, one from the 3D branch.
Fusing them per pixel would mix the two only where both exist, so the 3D-branch depth would refine only the SfM-observed pixels without affecting their neighbors.
We therefore employ spatial propagation~\cite{cheng2019learning}, which iteratively updates a dense depth map with learned, content-dependent neighbor weights to mix the two estimates adaptively across the entire image.

Specifically, we adapt the dynamic variant~\cite{lin2022dynamic} and propagate in linear depth, $D=\exp\zeta$.
On SfM-observed pixels, the confidence $\beta_i$ of Eq.~\eqref{eq:3d} blends the SfM and 3D-branch depths into an anchor, $a_i=(1-\beta_i)\,D^{\mathrm{sfm}}_i+\beta_i\,D^{\mathrm{3D}}_i$.
The 2D branch is equipped with a second decoder, which shares the encoder after the 2D--3D interaction and predicts the propagation parameters: affinities $w_{t,k}$ and offsets $\mathbf{o}_{t,k}$ over $K=5$ neighbors for each round $t$, and an anchoring weight $\eta_i$ set to zero where no SfM observation exists.
Starting from $D_i^{(0)} = D_i^{\mathrm{2D}}$, each of the $T$ rounds updates:

\begin{equation}
D_i^{(t)}(p) = (1-\eta_i)\sum_{k=1}^{K} w_{t,k}\, D_i^{(t-1)}(p+\mathbf{o}_{t,k}) + \eta_i\, a_i,
\end{equation}
where $\eta_i$, $a_i$, $w_{t,k}$, and $\mathbf{o}_{t,k}$ are taken at $p$, and $w_{t,\cdot}$ is normalized by a softmax over the $K$ neighbors.
After $T=6$ rounds, $D^{pred}_i = D_i^{(T)}$ gives the refined depth, which is back-projected with the SfM cameras to obtain the point map $P_i$.

\subsection{Training objectives}
\label{sec:method-training}

We supervise the outputs of the 2D branch, the 3D branch, and the propagation.
For the dense outputs, we use the confidence-weighted loss of~\cite{wang2024dust3r,chen2026ggpt} with the depth error,
\begin{equation}
  \mathcal{L}_{\mathrm{conf}}(D,c)=\frac{1}{|\Omega_i|}\sum_{p\in\Omega_i}
  c(p)\,\big|D(p)-D^{\mathrm{gt}}(p)\big|-\alpha\log c(p),
  \label{eq:loss-conf}
\end{equation}
where $\Omega_i$ is the set of pixels with valid ground truth, $c_i=1+\exp(\hat{c}_i)\ge1$ is the confidence obtained from the logit of Eq.~\eqref{eq:2d}, and $\alpha=0.2$ balances the regularizer.
The 3D branch is supervised only at the view's own SfM pixels with a plain L1 loss, $\mathcal{L}_{\mathrm{3D}}=\frac{1}{|\Omega_i\cap M_i|}\sum_{p\in\Omega_i\cap M_i}|D^{\mathrm{3D}}_i(p)-D^{\mathrm{gt}}_i(p)|$, without confidence.
The total loss is
\begin{equation}
  \mathcal{L}=\mathcal{L}_{\mathrm{conf}}(D^{pred}_i,c_i)+\lambda_{\mathrm{2D}}\,\mathcal{L}_{\mathrm{conf}}(D^{\mathrm{2D}}_i,c_i)+\lambda_{\mathrm{3D}}\,\mathcal{L}_{\mathrm{3D}},
  \label{eq:total}
\end{equation}
where the first term supervises the final output, and the second keeps a gradient on the 2D-branch depth from which the propagation starts.

%% file: figures/subfigure.tex
\begin{figure}[t]
    \centering
    \includegraphics[width=\columnwidth]{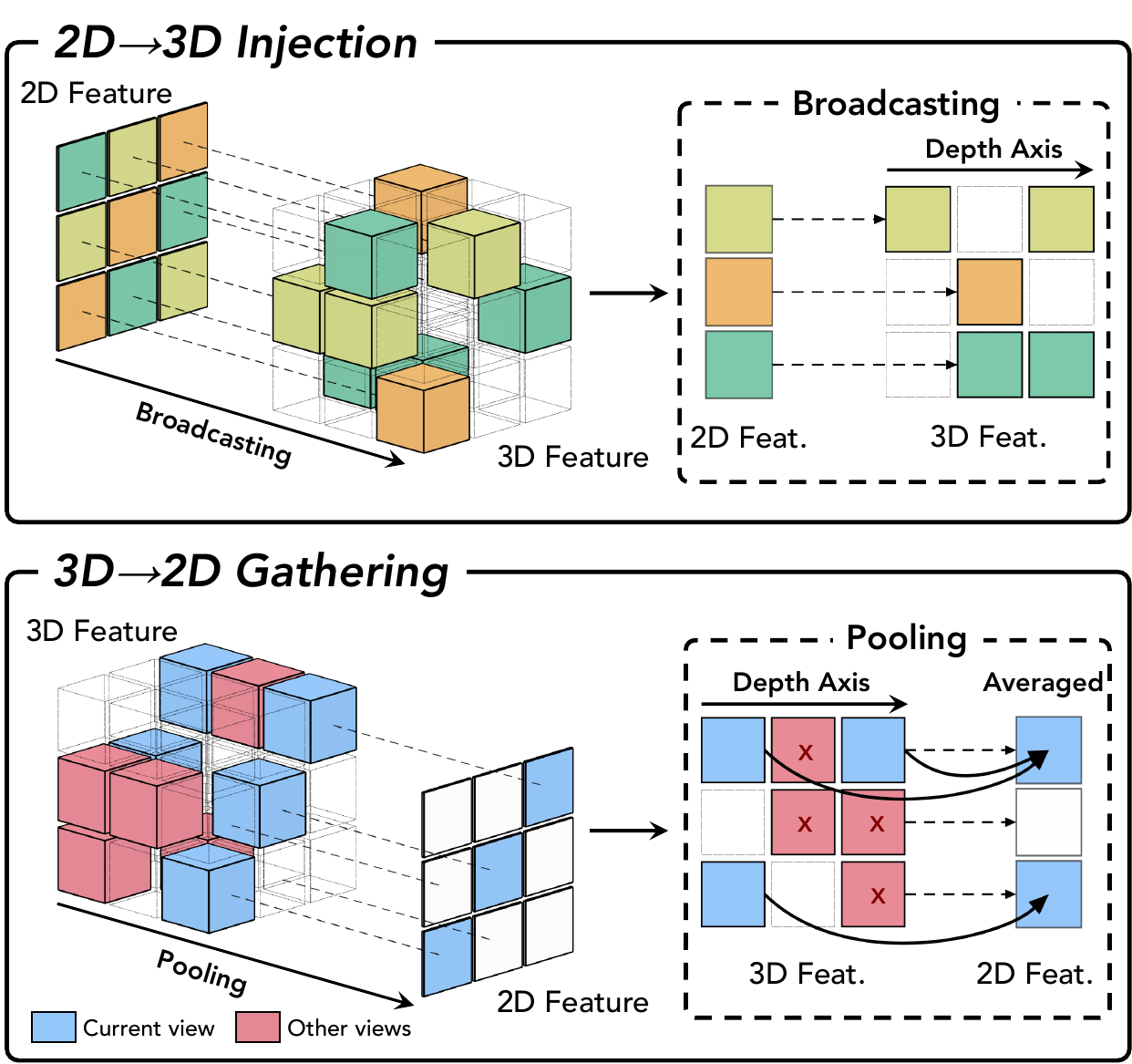}
    \caption{\textbf{Bidirectional 2D--3D interaction.}
     The 2D feature map of each view and the voxel features of its grid exchange information along pixel columns.
\emph{Top}: In 2D$\rightarrow$3D injection, the image feature of each pixel is broadcast to all occupied voxels in its column, providing the 3D branch with image context.
\emph{Bottom}: In 3D$\rightarrow$2D gathering, the voxel features in each column are average-pooled along the depth axis and fused into the image feature map, providing multi-view evidence back to the 2D branch.
Only voxels containing points of the current view (blue) are pooled, while those formed solely by points of other views (red) are excluded.} 
\vspace{-8mm}
    \label{fig:2d3dinteraction}
\end{figure}

%% file: sec/4_experiment.tex
\section{Experiments}

\subsection{Implementation details}

\paragraph{Inputs and architecture.}

For both training and evaluation, we obtain the initial reconstructions from a feed-forward backbone, either DA3~\cite{lin2025depth} or Pi3X~\cite{wang2026pi}, and the SfM pipeline of GGPT~\cite{chen2026ggpt} (Sec.~\ref{sec:method-formulation}); we report results for both versions, each trained separately. The 2D branch follows the U-Net of PacGDC~\cite{wang2025pacgdc}, the DPT head follows the architecture of the DA3 depth head, and the 3D branch follows the sparse U-Net of MoGe-3~\cite{moge3}.

\paragrapht{Training.}

We train on the training splits of ScanNet++~\cite{scannetpp}, TartanAir~v2~\cite{wang2020tartanair, ufm}, UnrealStereo4K~\cite{unrealstereo4k}, and BlendedMVS~\cite{yao2020blendedmvs}, all of which provide dense ground-truth depth and camera poses. Following MapAnything~\cite{keetha2026mapanything}, we sample multi-view sets via a covisibility-constrained random walk, with $4$--$16$ views for ScanNet++, $4$--$8$ views for TartanAir v2 and UnrealStereo4K, and $3$–$6$ for BlendedMVS. We train for $200$K iterations on $4$ NVIDIA RTX A6000 GPUs using AdamW~\cite{adamw} with a learning rate of $10^{-4}$ and weight decay $0.05$.
The learning rate is linearly warmed up over the first $2$K iterations and then cosine-decayed to $10^{-5}$. Further architectural and training details are in the supplementary material.

\subsection{Evaluation protocol}

\paragraph{Datasets.}
Following GGPT~\cite{chen2026ggpt}, we evaluate on ScanNet++~\cite{scannetpp}, ETH3D~\cite{eth3d_dataset}, and Tanks\&Temples~\cite{tandt}, each with 4-, 8-, and 16-view splits.
Since ScanNet++ is part of our training data, we refer to it as within-domain, and to ETH3D and Tanks\&Temples as cross-domain.
We further evaluate on two out-of-domain datasets whose appearance and geometry differ substantially from the training data: 4D-DRESS~\cite{4ddress}, rendered from high-resolution scans of clothed humans, and MV-dVRK~\cite{mvdvrk}, photorealistic renderings of robotic abdominal surgery.

\paragrapht{Metrics.} 
We evaluate point map accuracy with the area under the accuracy curve (AUC) at several error thresholds. We align each predicted point map to the metric ground truth with robust Umeyama alignment~\cite{umeyama} implemented in pycolmap~\cite{colmap}, and measure the Euclidean error between each predicted point and its ground-truth counterpart.

\paragrapht{Baselines.}

We compare against four groups of methods.
For fairness, all methods that take geometric priors receive the same SfM output as ours.
(1) \emph{Feed-forward models}: DA3~\cite{lin2025depth}, Pi3X (RGB only)~\cite{wang2026pi}, and VGGT-$\Omega$~\cite{wang2026vggt}, which take only images as input.
(2) \emph{Geometry-conditioned feed-forward models}: MapAnything~\cite{keetha2026mapanything} and Pi3X, conditioned on our SfM cameras and the per-view SfM depth $D^{\text{sfm}}_i$ (Sec.~\ref{sec:method-formulation}).
(3) \emph{Depth completion}: Murre~\cite{murre} and Any2Full~\cite{any2full}, which complete $D^{\text{sfm}}_i$ of each view independently; the results are back-projected with the SfM cameras into a global point map.
(4) \emph{Feed-forward reconstruction with multi-view geometry}: MASt3R-SfM~\cite{duisterhof2025mast3r} and BLASt3R~\cite{blast3r}, which run their own pipelines to jointly optimize cameras and dense depth using correspondences and depth priors, and GGPT~\cite{chen2026ggpt}, which refines the feed-forward point maps with the SfM points in 3D. We apply GGPT to the same DA3 and Pi3X initial reconstructions as ours.

\subsection{Results}

We summarize the main results below.
Additional qualitative examples and detailed quantitative results are provided in the supplementary material.

\begin{table*}[t!]
  \centering
  \small
  \setlength{\tabcolsep}{5pt}
  \renewcommand{\arraystretch}{1.1}
    \caption{\textbf{Multi-view 3D reconstruction on standard test sets.} We report AUC@1/5\,cm (\%\,$\uparrow$).
  ScanNet++ is \dsid{within-domain}, and ETH3D and Tanks\&Temples are \dscd{cross-domain}.
  The best and second-best results for each metric within each dataset are shown in \textbf{bold} and \underline{underline}, respectively.}
  \begin{tabular}{@{}l ccc ccc ccc@{}}
    \toprule
    & \multicolumn{3}{c}{\dsid{\textbf{ScanNet++}}~\cite{scannetpp}}
    & \multicolumn{3}{c}{\dscd{\textbf{ETH3D}}~\cite{eth3d_dataset}}
    & \multicolumn{3}{c}{\dscd{\textbf{T\&T}}~\cite{tandt}} \\
    \cmidrule(lr){2-4} \cmidrule(lr){5-7} \cmidrule(l){8-10}
    & 4 imgs & 8 imgs & 16 imgs & 4 imgs & 8 imgs & 16 imgs & 4 imgs & 8 imgs & 16 imgs \\
    \midrule
    \multicolumn{10}{@{}l}{\textit{Feed-forward}} \\
    Depth Anything 3~\cite{lin2025depth} & \res{1.6}{8.8} & \res{0.9}{7.1} & \res{0.6}{4.7} & \res{8.9}{31.9} & \res{8.0}{29.6} & \res{7.0}{27.1} & \res{1.1}{8.7} & \res{0.7}{7.1} & \res{0.5}{6.3} \\
    Pi3X (RGB only)~\cite{wang2026pi} & \res{14.7}{50.4} & \res{15.3}{53.9} & \res{14.3}{54.8} & \res{9.8}{34.8} & \res{6.1}{28.8} & \res{4.8}{26.4} & \res{5.1}{25.2} & \res{3.9}{23.5} & \res{3.4}{22.7} \\
    VGGT-$\Omega$~\cite{wang2026vggt} & \res{2.1}{16.3} & \res{1.3}{13.5} & \res{1.0}{11.9} & \res{4.1}{25.6} & \res{3.1}{22.2} & \res{2.2}{17.2} & \res{2.0}{16.0} & \res{1.5}{14.6} & \res{0.9}{9.9} \\
    \midrule
    \multicolumn{10}{@{}l}{\textit{Geometry-conditioned Feed-forward}} \\
    MapAnything~\cite{keetha2026mapanything} & \res{15.6}{43.9} & \res{13.3}{43.8} & \res{12.5}{42.0} & \res{6.8}{27.3} & \res{6.3}{25.0} & \res{5.3}{25.2} & \res{3.4}{18.7} & \res{2.5}{15.1} & \res{2.2}{14.7} \\
    Pi3X~\cite{wang2026pi} & \res{17.3}{52.1} & \res{18.2}{55.5} & \res{18.3}{58.3} & \res{15.6}{43.7} & \res{12.9}{40.0} & \res{10.2}{37.9} & \res{8.9}{33.6} & \res{8.1}{34.9} & \res{6.4}{30.7} \\
    \midrule
    \multicolumn{10}{@{}l}{\textit{Depth completion}} \\
    Any2Full~\cite{any2full} & \res{22.9}{50.0} & \res{28.6}{59.6} & \res{32.0}{65.1} & \res{24.8}{49.3} & \res{28.5}{55.4} & \res{30.4}{58.9} & \res{12.1}{34.4} & \res{13.1}{42.1} & \res{13.9}{42.7} \\
    Murre~\cite{murre} & \res{16.6}{44.1} & \res{21.3}{53.1} & \res{23.9}{57.4} & \res{13.0}{34.4} & \res{13.6}{35.4} & \res{15.2}{38.7} & \res{5.5}{22.2} & \res{6.6}{25.7} & \res{7.7}{28.2} \\
    \midrule
    \multicolumn{10}{@{}l}{\textit{Feed-forward w/ multi-view geometry}} \\
    MASt3R-SfM~\cite{duisterhof2025mast3r} & \res{11.5}{30.8} & \res{15.6}{41.0} & \res{18.9}{49.2} & \res{15.0}{37.4} & \res{17.5}{37.7} & \res{16.7}{39.0} & \res{14.9}{35.1} & \res{18.2}{41.6} & \res{17.7}{41.4} \\
    BLASt3R~\cite{blast3r} & \res{27.8}{57.1} & \res{34.5}{64.7} & \res{39.2}{67.5} & \res{11.7}{31.2} & \res{13.6}{34.9} & \res{14.2}{38.4} & \res{4.1}{19.1} & \res{5.2}{22.7} & \res{5.0}{22.8} \\
    GGPT (w/ DA3)~\cite{chen2026ggpt} & \res{21.6}{45.0} & \res{27.4}{56.9} & \res{31.8}{64.5} & \res{27.7}{56.2} & \res{28.4}{58.7} & \res{32.6}{64.4} & \res{14.9}{36.9} & \res{16.0}{46.1} & \res{16.0}{47.1} \\
    GGPT (w/ Pi3X)~\cite{chen2026ggpt} & \res{25.6}{56.7} & \res{29.5}{62.8} & \res{34.8}{67.8} & \res{31.7}{60.0} & \res{32.7}{62.0} & \res{39.4}{69.1} & \res{\underline{16.1}}{38.5} & \res{19.1}{49.7} & \res{19.7}{49.4} \\
    \midrule
    \textbf{\ours{} (w/ DA3) (Ours)} & \res{\underline{31.5}}{\underline{60.6}} & \res{\underline{37.5}}{\underline{68.8}} & \res{\underline{43.4}}{\underline{74.0}} & \res{\underline{42.7}}{\underline{67.3}} & \res{\textbf{46.4}}{\underline{69.9}} & \res{\textbf{53.6}}{\underline{76.1}} & \res{15.6}{\textbf{40.3}} & \res{\underline{20.5}}{\textbf{52.1}} & \res{\underline{21.8}}{\textbf{53.1}} \\
    \textbf{\ours{} (w/ Pi3X) (Ours)} & \res{\textbf{33.7}}{\textbf{65.6}} & \res{\textbf{39.6}}{\textbf{71.4}} & \res{\textbf{45.2}}{\textbf{75.7}} & \res{\textbf{43.1}}{\textbf{67.6}} & \res{\underline{46.3}}{\textbf{70.2}} & \res{\underline{53.3}}{\textbf{76.4}} & \res{\textbf{17.1}}{\underline{39.7}} & \res{\textbf{21.5}}{\underline{52.0}} & \res{\textbf{22.4}}{\underline{52.3}} \\
    \bottomrule
  \end{tabular}
  \vspace{-1mm}
  \label{tab:main}
\end{table*}
\paragrapht{Within- and cross-domain evaluation.}
Tab.~\ref{tab:main} compares \ours{} with the baselines on ScanNet++, ETH3D, and Tanks\&Temples.
\ours{} achieves the best accuracy across all benchmarks and, with either backbone, consistently outperforms GGPT applied to the same initial reconstructions.
Geometry-conditioned models such as Pi3X also benefit from the same SfM priors but remain far behind \ours{}.

\paragrapht{Out-of-domain evaluation.}
Tab.~\ref{tab:main_ood} reports results on 4D-DRESS and MV-dVRK, whose appearance and geometry differ substantially from the training data.
While the feed-forward backbones largely fail on these scenes, \ours{} achieves the best results on both datasets, with the largest margins at the strictest threshold.
VGGT-$\Omega$ is competitive at the looser threshold on 4D-DRESS, likely owing to human-centric training data~\cite{Black_CVPR_2023}, yet falls far behind \ours{} at the 1\,cm threshold.

\paragrapht{Qualitative comparison.}
Fig.~\ref{fig:main_qual_comp} compares point error maps of \ours{} and the baselines on all datasets. \ours{} keeps the accuracy of the SfM points while refining regions they do not cover, such as textureless surfaces and single-view regions, where the baselines retain large errors.

\begin{table}[t]
  \centering
  \small
  \setlength{\tabcolsep}{3pt}
    \caption{\textbf{\dsod{Out-of-domain} evaluation.} AUC (\%\,$\uparrow$) on 4D-DRESS and MV-dVRK, with thresholds in centimeters and millimeters, respectively. The best and second-best results for each metric within each dataset are shown in \textbf{bold} and \underline{underline}, respectively.}

  \begin{tabular}{@{}l cc cc@{}}
    \toprule
    & \multicolumn{2}{c}{\dsod{\textbf{4D-DRESS}}~\cite{4ddress}}
    & \multicolumn{2}{c}{\dsod{\textbf{MV-dVRK}}~\cite{mvdvrk}} \\
    \cmidrule(lr){2-3} \cmidrule(l){4-5}
    & @1\,cm & @5\,cm & @1\,mm & @5\,mm \\
    \midrule
    \multicolumn{5}{@{}l}{\textit{Feed-forward}} \\
    Depth Anything 3                          & 2.1 & 6.3 & 17.2 & 50.7 \\
    Pi3X (RGB only)                & 3.6 & 11.8 & 3.9 & 21.1 \\
    VGGT-$\Omega$                               & 42.0 & 81.6 & 17.7 & 54.0 \\
    \midrule
    \multicolumn{5}{@{}l}{\textit{Geometry-conditioned Feed-forward}} \\
    MapAnything                              & 7.2 & 29.8 & 32.2 & 69.1 \\ 
    Pi3X                                   & 45.7 & 76.9 & 33.8 & 65.2 \\ 
    \midrule
    \multicolumn{5}{@{}l}{\textit{Depth completion}} \\
    Any2Full                                 & 63.2 & 79.7 & 65.3 & \underline{84.1} \\
    Murre                                    & 48.5 & 74.1 & 59.5 & 81.7 \\
    \midrule
    \multicolumn{5}{@{}l}{\textit{Feed-forward w/ multi-view geometry}} \\
    MASt3R-SfM                  & 51.1 & 70.1 & 43.9 & 64.6 \\ 
    BLASt3R                     & 13.9 & 39.6 & 24.6 & 50.8 \\ 
    GGPT (w/ DA3)                & 50.7 & 71.2 & 67.0 & 83.1 \\
    GGPT (w/ Pi3X)              & 64.6 & 82.1 & 62.6 & 76.6 \\
    \midrule
    \textbf{\ours{} (w/ DA3) (Ours)}           & \underline{75.1} & \underline{83.6} & \textbf{72.3} & \textbf{87.4} \\
    \textbf{\ours{} (w/ Pi3X) (Ours)}         & \textbf{80.2} & \textbf{89.6} & \underline{68.9} & 83.7 \\
    \bottomrule
  \end{tabular}
  \label{tab:main_ood}

\end{table}
\input{figures/main_qual_comp}

\subsection{Ablation study}
\label{sec:ablation}

\begin{table}[t]
  \centering
  \small
  \setlength{\tabcolsep}{3pt}
    \caption{\textbf{Ablation study.} AUC (\%) at 1, 5, and 10\,cm on ScanNet++ and ETH3D, averaged over the 4-, 8-, and 16-view splits.
  Rows (a)--(c) add one component at a time, and the last two rows compare how the depth of the 3D branch reaches the output on top of (c).
  $^\dagger$Applied at test time only, without retraining.}
  \resizebox{\columnwidth}{!}{%
  \begin{tabular}{@{}l l cc@{}}
    \toprule
    & & ScanNet++ & ETH3D \\
    & & @1 / @5 / @10$\uparrow$ & @1 / @5 / @10$\uparrow$ \\
    \midrule
    & Feed-forward (DA3)               & \phantom{0}1.0 / \phantom{0}6.9 / 15.1 & \phantom{0}8.0 / 29.5 / 43.6 \\
    & GGPT (w/ DA3)                    & 27.0 / 55.5 / 67.5 & 29.5 / 59.8 / 71.0 \\
    \midrule
    (a) & 2D branch, geometry only     & 34.3 / 64.7 / 76.6 & 42.8 / 67.4 / 76.3 \\
    (b) & \;+ image features           & 35.4 / 66.3 / 77.9 & 44.7 / 68.8 / 77.2 \\
    (c) & \;+ 3D branch \& interaction & 36.7 / 67.2 / 78.4 & 46.2 / 69.7 / 77.9 \\
        & \;\;+ hard substitution$^\dagger$ & 37.1 / 67.5 / 78.6 & 46.7 / 70.3 / 78.5 \\
    \midrule
    (d) & \;\;+ Spatial fusion (\ours{})  & \textbf{37.5} / \textbf{67.8} / \textbf{78.9} & \textbf{47.6} / \textbf{71.1} / \textbf{79.1} \\
    \bottomrule
  \end{tabular}
  }
  \label{tab:ablation}
  \vspace{-3mm}
\end{table}

Tab.~\ref{tab:ablation} ablates \ours{} on ScanNet++ and ETH3D. Rows (a)--(c) add one component at a time, and the last two rows compare how the 3D-branch depth reaches the output. All variants use DA3 and share the same training setting. 

\paragrapht{Image-grid refinement.}
Refining the feed-forward depth on the image grid from the geometric channels of Eq.~\eqref{eq:geo} alone (a) already outperforms GGPT by a large margin, showing the benefit of refining on the image grid rather than as points in 3D. Adding the backbone features (b) further improves accuracy, as they provide geometric cues where the SfM depth alone is insufficient.

\paragrapht{Bidirectional interaction.}

Adding the 3D branch with bidirectional interaction (c) brings consistent gains on both datasets. Since the 3D-branch depth is used only for supervision in this variant, these gains come solely from the feature exchange, which provides the 2D branch with multi-view evidence beyond the view's own SfM depth.

\paragrapht{Spatial fusion.}

Substituting the 3D-branch depth on SfM-observed pixels at test time already helps, confirming that the 3D branch refines these pixels more accurately than the 2D branch.
However, hard substitution trusts every triangulated pixel equally and leaves its neighbors untouched.
Spatial fusion (d) instead anchors this depth with learned confidence and propagates it through learned affinities, yielding the best results.

\subsection{Efficiency}

Tab.~\ref{tab:efficiency} reports the inference time and peak GPU memory on ETH3D. Since all methods in the table receive the same SfM front-end output, we report only the time of each method itself; ours and Any2Full process the views in batches of four.
Among them, Refin3R is the fastest at every view count, 2.2$\times$ faster than Pi3X and 15$\times$ faster than GGPT with 16 views. Our main numbers exclude the DA3 encoder, which already runs to obtain the SfM priors; even including it, Refin3R remains the fastest.

\begin{table}[t]
\centering
\small
\setlength{\tabcolsep}{4pt}
\caption{\textbf{Efficiency.} Inference time and peak GPU memory on ETH3D, measured on an NVIDIA RTX A6000. The last row includes the DA3 encoder.}
\label{tab:efficiency}
\resizebox{\columnwidth}{!}{%
\begin{tabular}{@{}l rrr r@{}}
\toprule
\multirow{2}{*}{Method} & \multicolumn{3}{c}{Time per scene (s) $\downarrow$} & Mem.\ (GB) $\downarrow$ \\
\cmidrule(lr){2-4}\cmidrule(l){5-5}
 & 4 views & 8 views & 16 views & 16 views \\
\midrule
Any2Full    & 0.78 & 1.58 & 3.17  & \textbf{5.0} \\
MapAnything & 0.70 & 1.37 & 2.76  & 12.3 \\
Pi3X       & 0.46 & 0.88 & 1.88  & 10.9 \\
GGPT        & 4.36 & 7.20 & 12.92 & 6.2 \\
\midrule
\textbf{Ours} & \textbf{0.17} & \textbf{0.39} & \textbf{0.86} & \underline{5.1} \\
\textbf{Ours + DA3 Backbone} & \underline{0.35} & \underline{0.73} & \underline{1.58} & 9.8 \\
\bottomrule
\end{tabular}%
}
\end{table}

%% file: figures/main_qual_comp.tex
\begin{figure*}[t]
    \centering
    \includegraphics[width=\linewidth]{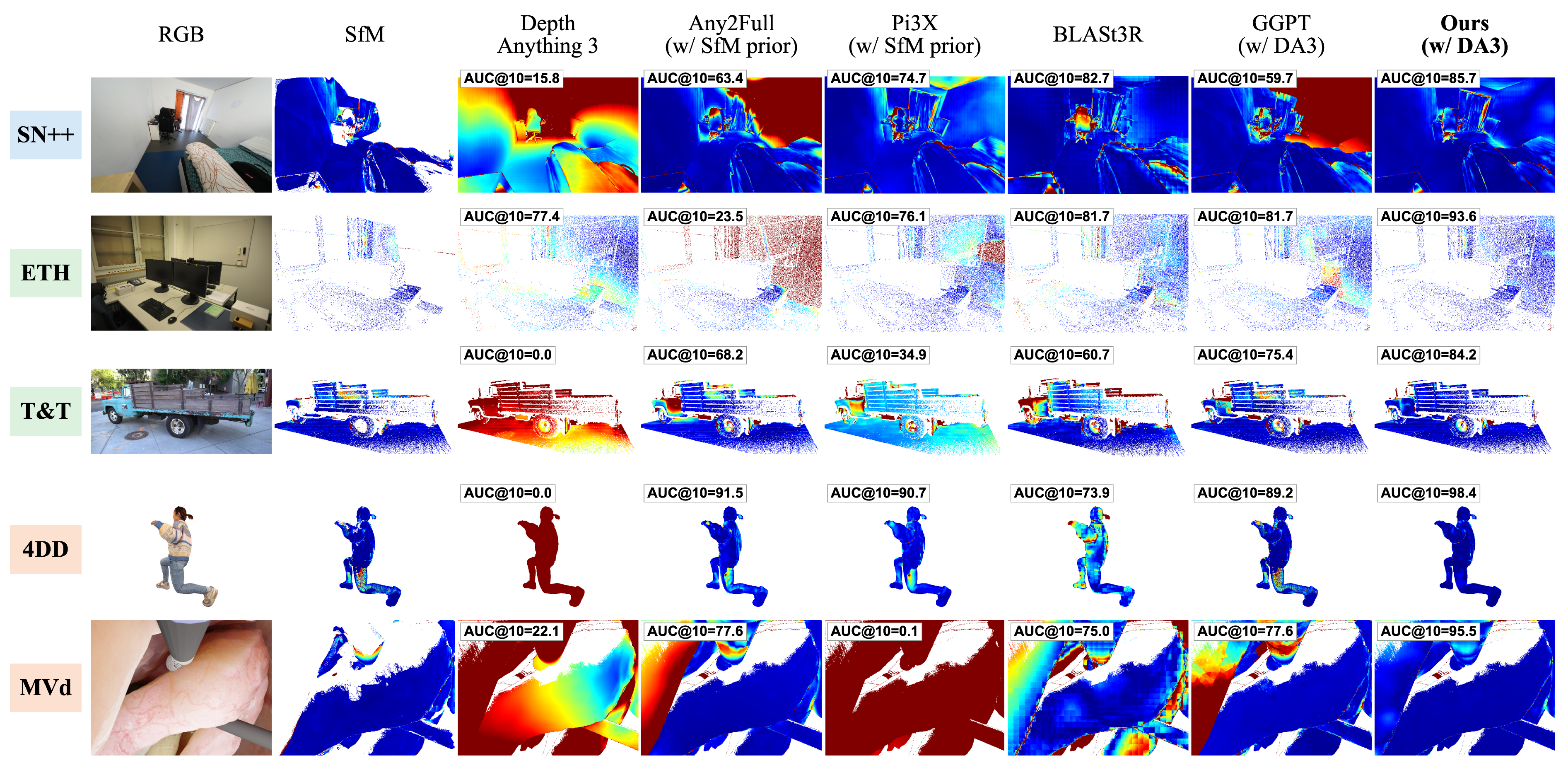}

    \caption{\textbf{Qualitative comparison.} 
    Each row shows one example from ScanNet++, ETH3D, Tanks\&Temples, 4D-DRESS, and MV-dVRK, with only one input view visualized. We compare a feed-forward model using only images (DA3), monocular depth completion (Any2Full), a geometry-conditioned model (Pi3X), and feed-forward methods with multi-view geometry (BLASt3R, GGPT), where Any2Full, Pi3X, and GGPT use the same SfM prior as ours. AUC@10 (cm; mm for MV-dVRK) is shown in each reconstruction.
    }
    \label{fig:main_qual_comp}
\end{figure*}

%% file: sec/5_conclusion.tex
\section{Conclusion}

We presented \ours{}, which refines feed-forward 3D reconstructions on the image grid of each view, guided by the SfM points of all views aggregated in 3D.
Its dual-branch architecture couples dense image features with multi-view SfM evidence through bidirectional interaction, and spatial fusion propagates the refined geometry across the whole image. 
\ours{} achieves state-of-the-art accuracy on standard and out-of-domain benchmarks, consistently outperforming feed-forward and geometry-guided methods while adding little overhead to the SfM front-end.

\newpage